\documentclass[10pt]{article} %
\usepackage[preprint]{tmlr}

\usepackage{amsmath,amsfonts,bm}

\def\eqref#1{equation~\ref{#1}}

\def\1{\bm{1}}

\DeclareMathAlphabet{\mathsfit}{\encodingdefault}{\sfdefault}{m}{sl}
\SetMathAlphabet{\mathsfit}{bold}{\encodingdefault}{\sfdefault}{bx}{n}

\usepackage{hyperref}
\usepackage{url}

\usepackage{soul}
\usepackage{url}
\usepackage[utf8]{inputenc}
\usepackage{booktabs}
\usepackage{color, colortbl}
\usepackage{subcaption,siunitx,booktabs}
\usepackage{multirow}
\usepackage{amssymb}
\usepackage{graphicx}
\usepackage{amsthm}
\usepackage{amsmath}

\newtheorem{lemma}{Lemma}
\newtheorem{definition}{Definition}
\newtheorem{assumption}{Assumption}
\newtheorem{proposition}{Proposition}

\newtheorem{remark}{Remark}

\usepackage{soul}
\usepackage{url}
\usepackage[utf8]{inputenc}
\usepackage{amsmath}
\usepackage{amsfonts}
\usepackage{booktabs}
\usepackage{color, colortbl}
\usepackage{multirow}
\usepackage{bm}
\newcommand{\mc}{\mathcal} 
\usepackage{mathtools}

\usepackage{thmtools}
\usepackage{thm-restate}

\title{Treatment Effect Estimation for Graph-Structured Targets}

\author{\name Shonosuke Harada \email fcb.sh1108@gmail.com\\
      \name Ryosuke Yoneda \\
      \name Hisashi Kashima\\
      \addr Kyoto University}

\def\month{MM}  %
\def\year{YYYY} %
\def\openreview{\url{https://openreview.net/forum?id=XXXX}} %

\begin{document}

\maketitle

\begin{abstract}
Treatment effect estimation, which helps understand the causality between treatment and outcome variable, is a central task in decision-making across various domains.
While most studies focus on treatment effect estimation on individual targets, in specific contexts, there is a necessity to comprehend the treatment effect on a group of targets, especially those that have relationships represented as a graph structure between them.
In such cases, the focus of treatment assignment is prone to depend on a particular node of the graph, such as the one with the highest degree, thus resulting in an observational bias from a small part of the entire graph. Whereas a bias tends to be caused by the small part, straightforward extensions of previous studies cannot provide efficient bias mitigation owing to the use of the entire graph information. In this study, we propose {\it Graph-target Treatment Effect Estimation (GraphTEE)}, a framework designed to estimate treatment effects specifically on graph-structured targets. GraphTEE aims to mitigate observational bias by focusing on confounding variable sets and consider a new regularization framework. Additionally, we provide a theoretical analysis on how GraphTEE performs better in terms of bias mitigation. Experiments on synthetic and semi-synthetic datasets demonstrate the effectiveness of our proposed method. 

\end{abstract}

\section{Introduction}

\begin{figure}[ht]
\begin{center}    \centerline{\includegraphics[width=0.95\columnwidth]{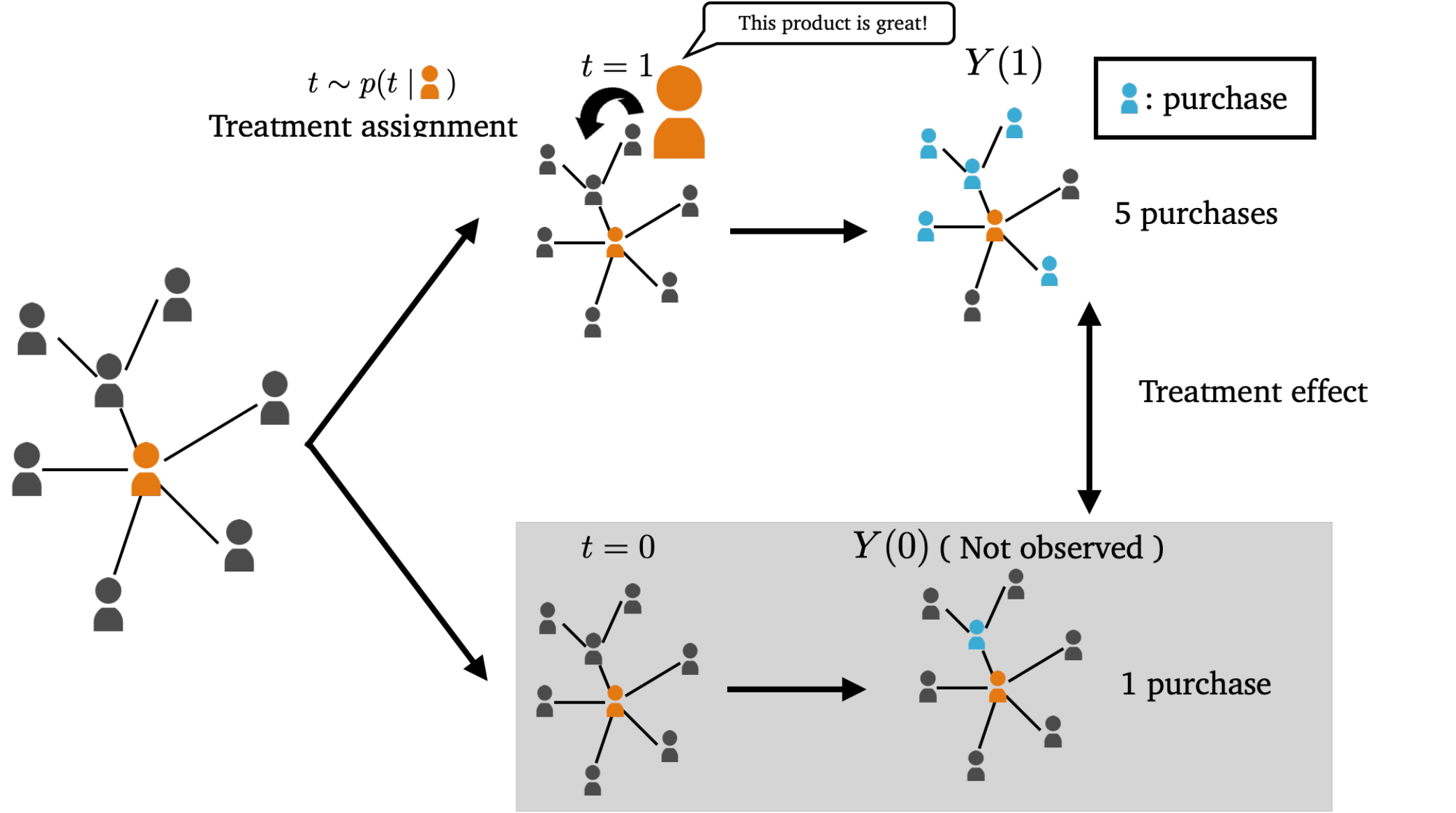}}
\caption{\small{An illustrative example of treatment effect estimation on graph-structured targets like a social network. The shaded outcome is a counterfactual outcome and never observed. For example, if a central user promotes a product, the entire graph may have positive impression on the product, i.g., the majority of users ($5$ users) may buy it; however, if no-advertisement is given, only $1$ user may have interests in the product and the others may not buy it. In this example, the treatment effect on this network is $4$. Our interest is to understand how interventions affect on entire graphs. }}
\label{fig:intro}
\end{center}
\end{figure}

The task of treatment effect estimation aims to understand the causal effect of a treatment or intervention on the outcome of a target object. It plays an crucial role in various fields such as advertisement~\citep{lewis2011here}, education~\citep{zhao2017estimating}, economic policy~\citep{lalonde1986evaluating}, healthcare~\citep{wood2008empirical}. 
For instance, in healthcare, treatment effect estimation is used to assess the safety and efficacy of developed medicines.

Most studies focus on treatment effect on a individual target; however, in some scenarios, treatment effect on graph-structured including set-structured targets need to be considered. For instance, an influencer, who has a large number of followers on a social network or social media because of their social position, could be involved in a form of marketing called influencer marketing~\citep{de2019influencer,lagree2018algorithms}. Figure~\ref{fig:intro} describes an example in a social network.
In influencer marketing, advertisement companies leverage influencers’ brands to promote their products, and this enables businesses to reach targeted individuals. 
Another practical application is group recommendation~\citep{cao2018attentive,felfernig2018group,sankar2020groupim}, which aims to recommend items for a group, rather than individuals. These groups of targets can be represented as graphs because they can capture complex relationships between entities and it becomes crucial to comprehend the treatment effect on graph-structured targets, rather than focusing solely on individual targets.
In these applications, understanding the treatment effect on graph-structured individuals becomes vital, not on individual.
There are mainly two fundamental challenges in treatment effect estimation, namely (i) observational bias and (ii) the counterfactual nature of observational data. Observational bias is a form of bias that occurs in the process of treatment assignment on target objects based on their covariates.
For example, the elderly are more likely to receive medical treatment than younger ones. Regarding the counterfactual nature of observational data, treatment effect is often defined as the difference between  outcomes of a graph with and without treatment; however, we can only observe one of the possible outcomes. Hence, studies have been focusing on understanding counterfactual outcomes.
Moreover, in treatment effect estimation for graph-structured targets, observational bias becomes a more severe technical difficulty to be addressed because a policymaker does not necessarily pay attention to all the nodes in a graph but, rather, to some specific nodes, such as a central node, often referred to as a hub node, that has a large number of connections compared to other nodes. In social networks, for example, an influencer who has a large number of followers is prone to attract a significant amount of attention for promoting products in influencer marketing. While the sizes of input graphs are large, the space for nodes which affect both treatment assignments and outcomes is relatively small. From the perspective of invariant learning on graph data, these features are well-known as invariant features that are essential for prediction~\citep{ma2020causal,wu2022dir}. Therefore, efficient mitigation of biases in graphs is a major technical challenge.
First, we provide theoretical analysis that shows how using entire node representations can introduce inefficient bias mitigation.
Then, we propose {\it Graph-target Treatment Effect Estimation (GraphTEE)}, a novel algorithm that efficiently mitigates bias based on theoretical analysis. GraphTEE consists of two major steps. First, we build a function that decomposes an input graph into necessary and non-necessary nodes for bias mitigation. Second, we apply an efficient bias mitigation regularizer, based on theoretical analysis.
In experiments using synthetic and semi-synthetic datasets, we validate the effectiveness of our proposed method.

The contributions of this study are summarized as: 
\begin{itemize}
\item As far as we know, this is the first study that focuses on a treatment effect estimation problem on graph-structured objects.
    \item We propose a novel method, GraphTEE, that mitigates biases more efficiently than the method of using entire graph nodes, particularly in large-sized graphs. We also provide a theoretical analysis that shows how existing approaches can suffer from inefficient bias mitigation.
    \item Through extensive experiments using synthetic and semi-synthetic datasets, we demonstrate the effectiveness of GraphTEE.
\end{itemize}

\section{Related Work}\label{sec:related}

\subsection{Treatment effect estimation}
Treatment effect estimation is a vital task that has been extensively studied in various fields, ranging from advertisement~\citep{lewis2011here}, economy~\citep{lalonde1986evaluating} and healthcare~\citep{eichler2016threshold,wood2008empirical} to education~\citep{zhao2017estimating}. 
One of the classic methods is the
matching method~\citep{rubin1973matching} which involves pairing target objects in a treatment group with target objects in a control group with similar characteristics, such as age, gender or health status. Propensity score matching~\citep{rosenbaum1983central,rosenbaum1985constructing} is a  class of the matching that uses a predictive model to estimate the probability of targets receiving treatments based on their covariates. The estimated probability, called propensity score, is used for matching targets. 
Deep learning-based methods have also been successfully applied to the treatment effect estimation problem~\citep{johansson2016learning,hassanpour2019learning,shalit2017estimating,NEURIPS2019_8fb5f8be}.
One of the promising methods is a CounterFactual Regression~(CFR)~\citep{shalit2017estimating} that is characterized by a regularizer designed to minimize the distribution discrepancy between treatment and control groups. CFR without any regularization is called Treatment-Agnostic Representation Network~(TARNet), which simply has representation sharing and outcome prediction layers for each group.
The key idea behind representation balancing is profoundly linked with domain adaptation analysis~\citep{blitzer2007learning,mansour2009domain,ganin2016domain}.
Shi et al.~\citep{NEURIPS2019_8fb5f8be} proposed a multi-task learning algorithm that learns representations while predicting the propensity score.
In recent times, the problems of treatment effect estimation using graph-structured treatments like molecules have also been introduced~\citep{harada2021graphite,kaddour2021causal}; however their goal is to estimate the effect of graph-structured treatments on individuals, rather than the treatment effect on graph-structured targets, and completely different from our focus.
Several studies have also considered treatment effect estimation on graph-structured data~\citep{guo2020learning,zheleva2021causal}; 
however, all these studies have considered treatment effect estimation on node level targets, rather than at the level of entire graphs.

\subsection{Representation learning on graph-structured data}
Because of the real world applications of graph-structured data, a substantial number of studies have been conducted in a wide range of fields, such as chemoinformatics~\citep{gilmer2017neural,duvenaud2015convolutional,harada2020dual}, social network data mining~\citep{fan2019graph,yanardag2015deep}, citation network analysis~\citep{feng2019hypergraph}, road network analysis~\citep{jepsen2019graph}, 
we briefly introduce some notable studies with regard to graph-structured data.
The main focus of studies on graph-structured data has been how to extract 'good' representations which preserve graph properties well for downstream tasks such as clustering, node and graph classification, and link prediction by a mapping function.
Graph kernels~\citep{kashima2003marginalized} or graph embedding approaches~\citep{perozzi2014deepwalk} aim to capture structural information. 
Besides, modern GNNs-based models are capable of fully making use of rich context information in graphs and have shown remarkable performances~\citep{hamilton2017inductive,morris2019weisfeiler}.
Numerous studies have also investigated mapping functions that handle set- or graph-structured data in terms of their  theoretical properties. Zaheer et al.~\citep{zaheer2017deep}~proposed DeepSets, which provided an extensive framework for neural networks to handle set-structured inputs. Xu et al.~\citep{xu2018powerful}~proposed Graph Isomorphism Network~(GIN)~and generalized the framework of DeepSets to graph-structured input. Corso et al.~\citep{corso2020principal} provided theoretical analysis on an injective function using GNNs given continuous space.

\vspace{0.5mm}
To the best of our knowledge, none of the previous studies has considered a treatment effect estimation problem on graph-structured data as targets in spite of the indispensability of graph-structured data in the real world. Our framework also generalizes to set-structured data and is able to handle this type of data.

\section{Problem statement}

Suppose we have a dataset $\mc{D}=\{G_i, t_i, Y^{t_i}_i\}^{N}_{i=1}\in\mc{G}\times\mc{T}\times\mc{Y}$, where $G_i=\{V_i, A_i\}_{i=1}^{N}$ is a graph and, $V_i\in\mc{V}, A_i$ represents a set of nodes and an adjacency matrix, respectively.
Set $V_i$ consists of $|V_i|$ nodes as $V_i=\{v_{i,0},..., v_{i,|V_i|}\}$ and each node $v_{i,j}\in R^{d}$ has $d$ dimensional covariates.
We assume the treatment space $\mc{T}$ is binary; i.e., $\mc{T}=\{0, 1\}$ and the outcome space is one-dimensional and continuous-valued; i.e., $\mc{Y}=R$. Thus, our goal is to learn a predictive function $g:\mc{G}\times{\mc{T}}\rightarrow\mc{Y}$. 
\begin{definition}
    The potential outcome $Y_i(t_i)$ of an graph $G_i$ is defined as the outcome of $G_i$ when the treatment $t_i$ is actually applied.
\end{definition}
We make assumptions for the identification of treatment effect.
\begin{assumption}
Positivity.
Positivity assumes that for $\forall G_i$, treatment assignment probability is larger than $0$ and less than $1$, i.e., any graph-structured target has positive probability of receiving all treatments as $1>p(t_{i}\mid G_i)>0$. 
\end{assumption}
\begin{assumption}
Stable unit treatment value assumption~(SUTVA). SUTVA assumes the potential outcomes of each target is independent of the treatment assignment of other targets.
\end{assumption}
\begin{assumption}
Strong ignorability. Strong ignorability assumes that the potential outcomes and treatment assignment are independent of each other for a given target graph; i.e., $Y(1), Y(0)\perp t_i \mid G_i$. We assume that input graphs include confounding variables.
\end{assumption}

\begin{definition}
\label{def:confounding_variable_sets}
A confounding node.
We call a node a confounding node when it has an effect on both treatment assignment and an outcome.
We define $V_i(c)$ a set of confounding nodes and $V_i(y)$ a set of non-confounding nodes in $G_i$.
\end{definition}

\begin{assumption}
    We assume the decomposition function that decomposes graphs, $V_i=V_i(c)\cup V_i(y)$, is the same function across different graphs
    and $\mid V_i(c)\mid\ll\mid V_i\mid$, unless $\mid V_i\mid=1$, i.e., a small part of a graph has an effect on treatment assignment.
\end{assumption}
This assumption enables us to learn a such function to decompose $V_i$ into $V_i(c)$ and $ V_i(y)$. We wish to emphasize that this assumption is reasonable, as particular nodes or sub-substructures play a crucial role in the context of invariant learning~\citep{ma2020causal,wu2022dir}.
We sometimes express graphs without $i$ for convenience.

\section{Methodology}
The proposed framework consists of two steps.
In the first step, our aim is to learn the decomposition function $\phi$ and find nodes or a sub-graph that have an effect on treatment assignment and an outcome. We achieve this task using GNNs and self-attention graph~(SAG)~pooling~\citep{lee2019self}. 
In the second step, using $\phi$ learned in the first step, we decompose the input graph into confounding node set $V(c)$ and non-confounding node set $V(y)$. 
This decomposition facilitates the model in mitigating bias with greater efficiency, as detailed in the theoretical analysis.

\subsection{Confounding nodes selecting step}
In the first step, we perform a propensity score learning using GNNs.
The goal of this step is to build a function that takes a graph as an input and returns confounding nodes and not-confounding nodes.
To select confounding nodes in a graph, we apply SAG-pooling technique~\citep{lee2019self}.
SAG-pooling aims to 
learn graph representation while giving larger weights on more important nodes that seem to have larger effects on a target value.
In this context, we need to specify a set of nodes that have effects on treatment assignment process.
We aim to find these nodes by selecting nodes based on their weights through applying SAG-pool in the final step of GNNs in $\phi$.
The prediction of treatment assignment is obtained as:
\begin{align}
g_t(G_i) = h_t\bigl(\sum^{\mid V_i \mid}_{i\in V_i}\mathrm{SAGP}(\Phi_t(G_i))\odot (\Phi_t(G_i))\bigr),
\end{align}
where $\mathrm{SAGP}$ denotes the SAG-pooling and returns weights of each node given node representations and adjacency matric, $\odot$ represents the Hadamard product, and $h_t$ is standard feedforward neural networks, respectively. 
The loss function $L_t$ for selecting confounding nodes are given as:
\begin{align}\label{eq:propensity_loss}
  \mc{L}_{t} = \sum_{i=1}^{N}-t_i\log \sigma(g_t(G_i)) -(1-t_i)\log(1-\sigma(g_t(G_i))),
\end{align}
where $\sigma$ denotes the sigmoid function. We use $\mathrm{SAGP}$ to select the confounding nodes as $\phi:\mathrm{top}(\mathrm{SAGP}(\Phi_t(G_i)), k)=\{V_i(c), V_i(y)\}$\label{eq:sag_k} where $\mathrm{top}$ is a function that return the top $\mathrm{k}\%$ nodes in terms of their weights in $V_i$ as $V_i(c)$ and the others as $V_i(y)$.

\begin{figure*}[tb]
  \begin{center}
  \centerline{\includegraphics[width=\linewidth]{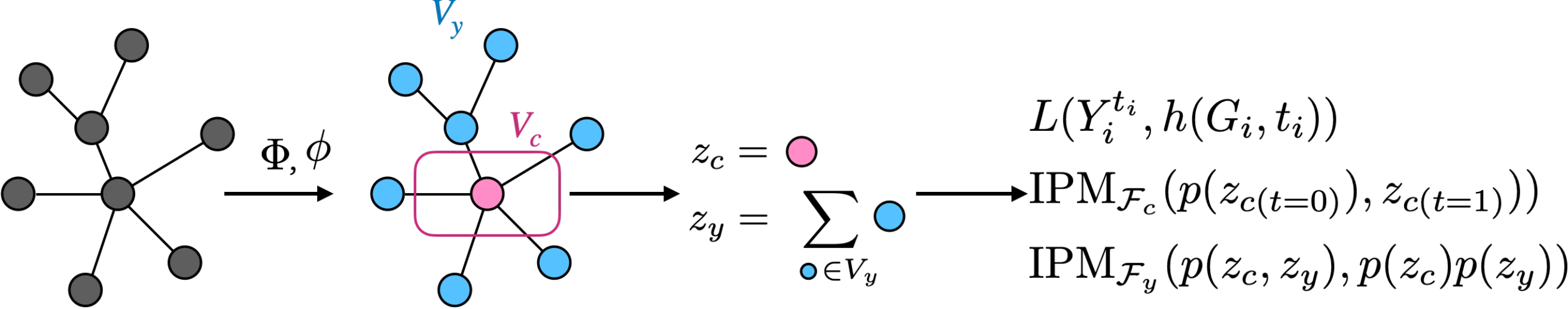}}
  \caption{Overview of the proposed algorithm. An input graph is first mapped into representations using GNNs $\Phi$ and then decomposed into two types of nodes using $\phi$:(i) confounding and (ii) other nodes. Outcome prediction is conducted using the entire nodes' representations with TARNet-based feedforward neural networks as Eq.~(\ref{eq:supervised_loss}). Representations of nodes of each type are used for bias mitigation as Eq.~(\ref{eq:ipm_loss}).}
  \label{fig:model}
  \end{center}
\end{figure*}

\subsection{Outcome prediction step}
In the second step of the proposed method, we implement a methodology for treatment effect estimation. Let $\Phi:\mc{G}\rightarrow \mc{Z}$ where $z\in\mc{Z}$ is a space of sets of representations, be a mapping function and $h:\mc{Z}\times\mc{T}\rightarrow\mc{R}$ be a hypothesis. 
We employ GNNs as the mapping function $\Phi$ for an input graph.
As discussed in the following section, several conditions have to be satisfied to fully use some theoretical benefits.
For instance, it is imperative for the GNNs to have a one-to-one mapping between input graphs and their representations, although neural networks that do not necessarily satisfy one-to-one mapping functions empirically can show satisfactory performances.
Recent theoretical analysis provided remarkable insights into the  theoretical properties of GNNs~\citep{xu2018powerful,zaheer2017deep}.
Particularly, we utilize a GIN-based GNN~\citep{xu2018powerful} to map graph-structured targets into representations $z\in\mc{Z}$. After performing the mapping, we obtain representations of each node for a input graph.
As a predictive model, we employ a TARNet-based model~\citep{shalit2017estimating}, which takes the graph representation and treatment indicator as inputs.
The supervised loss $\mc{L}_{\text{supervised}}$ is given as:
\begin{align}\label{eq:supervised_loss}
\mc{L}_{\text{supervised}}=\sum_{i=0}^{N} L(Y^{t_i}_i, g(G_i, t_i)),
\end{align}
where $g(G_i, t_i) \coloneqq h(\Phi(G_i), t_i)$. Due to the observational problem, the supervised loss itself does not give desirable results for the test distribution, necessitating the implementation of a bias mitigation technique.
We introduce a popular metric between probability distributions, the {\it integral probability metric~(IPM)} for bias mitigation.
The technique of IPM has been extensively utilized to measure the dissimilarity between probability distributions~\citep{muller1997integral,sriperumbudur2009integral,sriperumbudur2012empirical}.
\begin{definition}
The IPM for two probability distributions $p,q$ over $\mc{X}$ and for a function family $\mc{F}$ is defined as:
\begin{align}
    \text{IPM}_{\mc{F}}(p, q)=\underset{f\in\mc{F}}{\sup} \Bigl|\int_{\mc{X}} f(x)(p(x)-q(x))dx\Bigr|.
\end{align}
\end{definition}
Depending on the choice of function family $\mc{F}$, the IPM can be reduced to a widely recognized and utilized distance between two distributions. For example, the IPM yields the Wasserstein distance by choosing $\mc{F}$ as a $1$-Lipschitz function.
However, as we will discuss in the theoretical analysis,
using entire node representations in a graph makes empirical IPM computation inefficient.
In order to avoid such a situation, we apply the confounding nodes select function $\phi:\mc{G}\rightarrow \mc{V}, \mc{V}$, which is built in the first step, to decompose the sets of nodes into necessary and not necessary nodes for bias mitigation. 
We compute the two IPMs, i.e., $\text{IPM}_{\mc{F}}(p(z_{c(t=0)}), p(z_{c(t=1)}))$ and $\text{IPM}(p(z_c, z_y), p(z_c)p(z_c))$, where $z_c\coloneqq\{z\in z_j\mid j\in V(c)\}\in\mc{Z}_c$, $z_y\coloneqq\{z\in z_j\mid j\in V(y)\}\in\mc{Z}_y$, and $\mc{Z}=\mc{Z}_c\times \mc{Z}_y$. The first term aims to mitigate selection bias and the second one tries to decrease the dependency between $z_c$ and $z_y$.
Using these terms, the bias mitigation regularizer loss $\mc{L}_{\text{reg}}$ is given as:
\begin{align}\label{eq:ipm_loss}
\mc{L}_{\text{reg}} &= \text{IPM}_{\mc{F}_c}(p(z_{c(t=0)}), p( z_{c(t=1)})) \\&+ \text{IPM}_{\mc{F}_y}(p(z_c, z_y), p(z_c)p(z_c)).
\end{align}
We use the sum function as the aggregation function of nodes.
Note that the sum function ensures that $z_c+z_y=z=\Phi(G)$ and the mapping function has one-to-one mapping property~\citep{xu2018powerful,zaheer2017deep}.
Finally, the loss function to be minimized in the second step is given as:
\begin{align}\label{eq:proposed_loss}
\mc{L} = \mc{L}_{\text{supervised}} + \lambda\cdot\mc{L}_{\mathrm{reg}},
\end{align}
where $\lambda$ is a hyper-parameter that controls the strength of regularization. 
We describe the whole training process in Figure~\ref{fig:model}.
As we discuss in the experimental section,
it is not trivial to find good $\lambda$ because we can have access to only skewed observational data. 
We apply an existing approach to select a hyper-parameter. 
We optimize the loss function~(\ref{eq:proposed_loss})~using Adam~\citep{kingma2014adam} in a mini-batch manner.
Note that at the prediction phase, the selection procedure of confounding nodes is not necessary and straightforward GNNs are used to predict outcomes.

\section{Theoretical analysis}\label{sec:theoretical}
In this section, we provide a theoretical analysis with regard to our approach.
We define the expected the risk $\epsilon_{\text{train}}$ and $\epsilon_{\text{test}}$ using data from the train data distribution $p(G, t)$ and the test data distribution $p(G)p(t)$ as follows:
\begin{definition}
The expected risks on the train data distribution and the test data distribution are defined as:
\begin{align}
     \epsilon_{\mathrm{train}}(h, \Phi)=\mathbb{E}_{(G,t)\sim p(G,t)}[L(h(\Phi(G), t),   Y^{t})],\\
     \epsilon_{\mathrm{test}}(h ,\Phi)=\mathbb{E}_{(G,t)\sim p(G)p(t)}[L(h(\Phi(G), t),   Y^{t})],
\end{align}
where $L$ is a loss function, such as squared loss.
\end{definition}
Note that we assume the test data distribution we aim to evaluate our method on is $p(G, t)\coloneq p(G)p(t)$, i.e., a treatment assignment is randomized.
Empirically, we approximate the IPM using finite samples. For example, within deep learning-based framework, IPM is typically computed in a mini-batch manner.
We refer to $\text{IPM}_{\mc{F}}(\hat{p}, \hat{q})$ as such an empirical estimate of the IPM. 
Sriperumbudur et al.~\citep{sriperumbudur2009integral} analyzed the empirical risk of IPMs.
\begin{proposition}\label{proposition:ipm_error}~\citep{sriperumbudur2009integral}
Given $m$, $n$ samples from $p,q$ respectively and for any $\mc{F}$ such that $\nu\coloneqq\underset{{x\in\mc{X}}}{\sup}f(x)<\infty$, we have with a probability at least $1-\delta$, 
  \begin{align}
  &\Bigl|\text{IPM}_{\mc{F}}(p,q)-{\text{IPM}}_{\mc{F}}(\hat{p},\hat{q})\Bigr|\leq 2R_{m}(\mc{F})+2R_{n}(\mc{F})\nonumber\\&+\sqrt{18\nu^{2}\log\frac{4}{\delta}}(\frac{1}{\sqrt{m}}+\frac{1}{\sqrt{n}}),
  \end{align}
where $R_m(\mc{F})\coloneqq\mathbb{E}\underset{f\in\mc{F}}{\sup}\Bigl| \frac{1}{m}\Sigma^{m}_{i=1}\rho_if(x_i) \Bigr|$ and $\{\rho_{i}\}_{i=1}^{m}$ is the Rademacher complexity and  
independent Rademacher random variables that take $1$ or $-1$ at the probability $0.5$, respectively.
\end{proposition}
{\proof See the original paper~\citep{sriperumbudur2009integral}.}
The IPM has been successfully applied in recent treatment effect estimation methods that were based on representation learning to mitigate the selection bias~\citep{shalit2017estimating,johansson2016learning}.
Combining Proposition~\ref{proposition:ipm_error} with the expected risks, we can detive the following bound in a similar manner as the previous study~\citep{johansson2018learning}.
\begin{restatable}[]{proposition}{baselineproposition}
\label{proposition:cfr_ipm_error}
  Let $\Phi:\mc{G}\rightarrow\mc{Z}$ be a one-to-one mapping function. 
Let $h\colon\mc{Z}\times\{0,1\}\rightarrow\mc{Y}$ be a hypothesis.
Let $\mc{F}$ be a family of functions and $f\in\mc{F}\colon\mc{Z}\rightarrow\mc{Y}$.   
Assume there exists a constant ${B_\Phi}$ such that $\frac{1}{B_\Phi}\int_{\mc{Y}} L(Y^{t}, h(\Phi(G), t))p(Y^{t}\mid G)p(G, t)d{Y^{t}}\in\mc{F}$. Supposing $f=h(z)\in \mc{F}$, we have with a probability at least $1-\delta$,
  \begin{align}
    &\epsilon_{\mathrm{test}}(h, \Phi) \leq \epsilon_{\mathrm{train}}(h, \Phi)+ B_{\Phi}\cdot{\text{IPM}_{\mc{F}}(\hat{p}({z_{(t=0)}}),\hat{p}((z_{(t=1)})))}\\& + 2R_{m}(\mc{F})+2R_{n}(\mc{F})+\sqrt{18\nu^{2}\log\frac{4}{\delta}}(\frac{1}{\sqrt{m}}+\frac{1}{\sqrt{n}}).
  \end{align}
\end{restatable}
  {\proof Appendix.}

This regularization still provides a theoretically correct upper bound for counterfactual outcomes; however, if we naively apply this approach to a graph-structured targets setting, this regularization computes the IPM using all the node representations in the input sets and cannot avoid an inefficient computation. 
Based on the assumption that a small part of an input graph has an effect on the treatment assignment, rather than on the entire graph; i.e., confounding nodes account for small part of large graphs, we aim to establish a more efficient way to mitigate bias.

When each variable and node is independent of each other, we only need to use a confounding node for bias mitigation.
However, the technical difficulty when we focus on graph-structured targets is confounding nodes and other structures are basically relevant and dependent on each other. 
Therefore, we try to achieve efficient bias mitigation by focusing only on the confounding nodes. 
If there are no relationship between $z_c$ and $z_y$, i.e., $p(z)=p(z_c)p(z_y)$, only minimizing $\text{IPM}_{\mc{F}}(p({z_{c(t=0)}}), p({z_{c(t=1)}}))$ is sufficient to mitigate bias; however, due to the dependency between $z_c$ and $z_y$, i.e., $p(z_c,z_y)=p(z_c\mid z_y)p(z_y)$, just minimizing the IPM with regard to the confounding node representation is insufficient and we need to remove the dependency by minimizing  $\text{IPM}_{\mc{F}_c}(p({z_c}, z_y), p(z_c)p(z_y))$.
We give a theoretical analysis on the proposed method between this approach and the original regularization $\text{IPM}(p({z_{(t=0)}}), p({z_{(t=1)}}))$.

\begin{restatable}[]{theorem}{maintheorem}
\label{theorem:main}
Suppose we have an error as $\text{IPM}(p(t)p({z_c}\mid{z_y})p(z_y), p(t)p(z_c)p(z_y))=\Delta.$
We have with probability at least $1-\delta$,
\begin{align}
&\epsilon_{\mathrm{test}}(h, \Phi) \leq \epsilon_{\mathrm{train}}(h, \Phi) + {B_\Phi}\cdot\bigl\lparen{\text{IPM}}_{\mc{F}_c}(\hat{p}(z_{c(t=0)}),\hat{p}(z_{c(t=1)}))
 + {\text{IPM}}_{\mc{F}_y}(\hat{p}(z_c, z_y), \hat{p}(z_c)p(z_y))\bigr\rparen\nonumber
\\& +  2R_{m}(\mc{F}_c) +2R_{n}(\mc{F}_c)  
 + \sqrt{18C_{\nu}^{2}\log \frac{8}{\delta}}(\frac{1}{\sqrt{m}}+\frac{1}{\sqrt{n}})
  +4R_{m+n}(\mc{F}_y) + 
2\sqrt{ \frac{18{\nu}^{2}}{m+n} \log \frac{8}{\delta}}+ 2\Delta.
  \label{eq:main_regularization}
\end{align}
\end{restatable}
{\proof Appendix.}\\
\begin{remark}
The main benefit of the proposed method (Theorem~\ref{theorem:main}) comes from the analysis that the empirical error for the IPM terms can become smaller than the existing approach (Proposition~\ref{proposition:cfr_ipm_error}) mainly because the constant $C_c$ is smaller than 1 and using $m+n$ samples when we can minimize $\Delta$ enough. Hence, our regularization enables us to compute the regularization more efficiently than the existing one, particularly when $\mc{Z}_c$ is small compared to $\mc{Z}$.
Note that we cannot necessarily claim the upper bound of Theorem~\ref{theorem:main} is always tighter 
than the existing one~(Proposition~\ref{proposition:cfr_ipm_error}); 
however, we expect that space of set of confounding nodes representation $\mc{Z}_c$ is sufficiently smaller than the entire representation space to give computationally efficient bias mitigation, 
and moreover, by using both of them, with probability at least $1-2\delta$, we can provide a tighter bound. We demonstrate the effectiveness of this regularization empirically. 
\end{remark}

\section{Experiments}\label{sec:experiments}

\subsection{Datasets}

\noindent
{\bf Synthetic dataset}:
We generated a synthetic graph dataset using NetworkX package~\citep{hagberg2008exploring}\footnote{\url{https://networkx.org/}}.
Moreover, we generate random graphs following the Barab{\'a}si--Albert algorithm~\citep{barabasi1999emergence}.
Each node has $d$-dimensional covariates $v_{ij}\in{R^{d}}\sim\mathcal{N}(0, \mathrm{Degree(i, j)})$, where $\mathrm{Degree(i, j)}$ is a degree of node $j$ in $G_i$.
We assume treatment is assigned by a node who has the highest degree in a graph as $t_j\sim\text{Bern}(\alpha \sum_{d=1}^{d}\text{HighestDegree}(G_i))$, where $\text{Bern}$ represents the Bernoulli distribution, $\text{HighestDegree}(G_i): \mc{G}\rightarrow R^{d}$ is a function that returns covariates of node who has the highest degree, and $\alpha$ is a constant value that controls the strength of observational bias. If there are more than one node which has the highest degree, $\text{HighestDegree}$ returns the covariates of node whose sum of the covariates are the largest.
We use widely employed GNNs in generating outcomes as $\mathrm{GNNs_0}$ and $\mathrm{GNNs_1}$~\citep{morris2019weisfeiler} which are defined as:
\begin{align}
    &\{z_{i0,}(1)\ldots z_{i,|V_i|}(1)\}=\mathrm{gnn}^{w}_1(V_i, A_i),\\ &\mathrm{where}\ z_{ij}(1)= W^{w}_1(1)v_{ij} + W^{w}_2(1)\sum_{k\in\mc{N}_{A_i}(j)}v_{ik},\\
  & z_ij(2)=\mathrm{gnn}^{w}_2(\{z_{i0,}(1)\ldots z_{i,|V_i|}(1)\}, A_i, j)\\&\qquad\hspace{-1.5mm}\quad =W^{w}_1(2)z_{ij}(1) + W^{w}_2(2)\sum_{k\in\mc{N}_{A_i}(j)}z_{ik}(1),\\
    &\mathrm{GNNs}_{w}(G_i=(V_i, A_i), j)\coloneqq\mathrm{gnn^{w}_2}(\mathrm{gnn^{w}_1}(V_i, A_i)), j),
\end{align}
where $\mc{N}_{A_i}(j)$ represents a set of neighbors of $j$ in $A_i$.
Outcomes are generated as $Y_i^{0} = \frac{1}{d}\frac{1}{|V_i|}C_s\sum^{|V_i|}_{j}{z^{0}_{ij}}+ \tau, Y_i^{1} = \frac{1}{d}\frac{1}{|V_i|}C_s\sum^{|V_i|}_{j} {z^{1}_{ij}} + \tau$
where $z^{0}_{ij}=\text{GNNs}_{0}(G_i), z^{1}_{ij}=\text{GNNs}_{1}(G_i)$, and $C_s$ is a scaling parameter, and we set $C_s$ as $1$ for the synthetic dataset. $\tau$ is a noise sampled from $\mc{N}(0, 0.01)$. 
We set parameters of $\mathrm{GNNs}_0$; i.e., $W^{0}_{1}(1),\ldots W^{2}(2)$, using random variables sampled from $\mc{U}(-3, 3)$ and similarly $\mathrm{GNNs}_1$ from $\mc{U}(0, 3)$ so that graphs whose central nodes are larger values tend to have larger treatment effect and vice versa. $\mc{U}$ denotes a uniform distribution.
Note that when we set $\alpha$ as $0$, the treatment assignment reduces to random assignment and unless otherwise stated, we report experimental results when we set $\alpha$ as $0.5$.
This setting reflects a situation in which treatment is assigned only by a small part of an entire graph on which people are prone to focus.
We set each node as $20$-dimensional covariates and $2{,}000$ graphs are used in one experiment.

\vspace{0.75mm}
\noindent
{\bf Reddit dataset}~\citep{yanardag2015deep}:
We use Reddit binary dataset~\citep{yanardag2015deep} for a semi-synthetic dataset. In Reddit dataset, each graph represents an online discussion thread. Each node represents a user and edges are given between two nodes if at least one of them correspond to the other comments.
The original dataset is used for a binary classification problem to classify whether a graph is a question/answer-based thread or a discussion-based thread. 
In this paper, we make use of graphs included in the dataset as a real world data and generate a tuple of covariates, treatments and outcomes in the same manner as the synthetic data. we set $C_s$ as $0.5$ for the synthetic dataset.
We also set each node as $20$-dimensional covariates.
We remove graphs which have more than $500$ nodes for computational convenience and have $1{,}617$ graphs.

\begin{table*}[t]
    \caption{Performance comparisons on the synthetic and Reddit datasets in terms of $\sqrt{\epsilon_\text{PEHE}}$ and ${\epsilon_\text{ATE}}$. We conduct experiments $20$ times for different seeds, and show the average results and standard errors. We set the number of node as $100$ for the synthetic dataset. The bold results indicate the best results in terms of average, and $^\dagger$ and $^\ddagger$ indicate the proposed method perform e statistically significantly better than the baseline methods by the paired t-test with $p < 0.05$ and $p < 0.01$. Lower values are better. }\label{table:results}
\centering
\begin{tabular}{ccccccccc}
\toprule[2pt]
                & \multicolumn{2}{c}{ \bf{Synthetic} }        &              &  \multicolumn{2}{c}{ \bf{Reddit} } \\[3pt] \cmidrule{1-6}
Method                       & $\sqrt{\epsilon_{\text{PEHE}}}$ & $\epsilon_{\text{ATE}}$ & \multicolumn{1}{l}{} &  $\sqrt{\epsilon_{\text{PEHE}}}$ &  $\epsilon_{\text{ATE}}$  \\[3pt] \cmidrule{1-6}
\multicolumn{1}{c}{Mean }    &        $^\ddagger65.477_{\pm{7.447}}$       &       $^\ddagger29.918_{\pm{4.834}}$         && $^\ddagger 64.458_{\pm 5.347}$   & $^\ddagger12.786_{\pm 2.425}$  \\[2pt]
{DeepSets}    &        $^\ddagger 43.372_{\pm{1.292}}$       &       $^\ddagger17.415_{\pm{0.938}}$         && $^\ddagger 34.586_{\pm 1.316}$   & $^\ddagger7.239_{\pm 0.405}$  \\[2pt]
{DeepSets+CFR}    &        $^\ddagger 43.905_{\pm{1.174}}$       &       $^\ddagger18.917_{\pm{0.893}}$         && $^\ddagger 34.692_{\pm 1.610}$   & $^\ddagger6.672_{\pm 1.187}$  \\[2pt]
 GNN  &        $^\ddagger40.481_{\pm{1.706}}$       &       $^\ddagger12.248_{\pm{1.605}}$        &  &       $^\ddagger28.105_{\pm{4.738}}$        &       $^\dagger4.419_{\pm{0.848}}$        \\[2pt]
 GNN+CFR  &        $^\ddagger16.703_{\pm{0.637}}$       &       $4.556_{\pm{0.541}}$        &  &       $^\dagger 16.212_{\pm{0.628}}$        &       $2.523_{\pm{0.321}}$        \\[2pt]
GraphTEE  &        $\bf 15.644_{\pm{0.644}}$       &       $\bf 4.137_{\pm{0.519}}$        &  &       $\bf 15.317_{\pm{0.633}}$        &       $\bf 2.331_{\pm{0.352}}$        \\
\bottomrule[2pt]
    \end{tabular}
\end{table*}

\subsection{Experimental settings}
We compare the proposed method with the following baselines.
Note that due to the nature of input data, popular methods such as the TARNet or CFR~\citep{shalit2017estimating} cannot be
applied directly to this experiment in the original form and hence, none of previous studies have handled treatment effect estimation on graph-structured targets. Therefore we extend these methods~\citep{shalit2017estimating,zaheer2017deep} to this setting by replacing the neural networks with DeepSets or GNNs because they are unable to handle graph-structured data without modifications to the model architectures.
(i) {\it Mean} predicts the outcome using the mean of train dataset.
(ii) {\it DeepSets}~\citep{zaheer2017deep} predicts the outcome by combining DeepSets and TARNet.
(iii) {\it DeepSets-CFR} predicts the outcome by the combination of DeepSets and TARNet with the Wasserstein regularization as the IPM for bias mitigation.
(iv) {\it GNNs} predicts the outcomes using GNNs without any regularization.
(v) {\it GNNs-CFR} predicts the outcomes using GNNs with the Wasserstein regularization as the IPM for bias mitigation.
We split the dataset into $20\%$, $20\%$, and $60\%$ for train, validation, and test datasets, respectively.
Note that we particularly focus on a relatively small number of data because the error of bias mitigation becomes severer.
As a evaluation metric, we employ $\sqrt{\epsilon_{\text{PEHE}}}$ that measures regression error in terms of CATE and is defined as $\sqrt{\epsilon_{\text{PEHE}}} =  \frac{1}{N_\mathrm{test}}\sum^{N}_{i}((Y^{1}_i - Y^{0}_i) - (\hat{Y}^{1}_i-\hat{Y}^{0}_i))^{2},$
where $N_\mathrm{test}$ represents the number of samples in test data and $\hat{Y}^{t}_i$ is a predicted outcome given as $\hat{Y}^{t}=g(G_i, t)$.
We also employ $\epsilon_{\text{ATE}}$ that evaluates performances in terms of ATE defined as $\epsilon_{\text{ATE}} = \Bigl| \frac{1}{N_{\mathrm{test}}}\sum^{N}_{i=1}(Y^{1}_i - Y^{0}_i) -\frac{1}{N_{\mathrm{test}}}\sum^{N}_{i=1}(\hat{Y}^{1}_i-\hat{Y}^{0}_i)\Bigr|$. 
The regularization hyper-parameter $\lambda$ is selected from $\{ 10^{-3},\  10^{-2}, \ldots 10^2,\ 10^{3} \}$
based on the loss for the validation dataset.
We set the $k$ in selecting the confounding node in 
the SAG-pooling as $10$; i.e., $10\%$ nodes in each graph are treated as the confounding nodes.
For the computation of Wasserstein distance as the IPM regularization, we employ Sinkhorn's algorithm~\citep{cuturi2013sinkhorn}.
We use the ${\it elu}$ function as an activation function in all the models.
We set the number of layers to $3$ and the dimension of representations as $32$ through the experiments.

   \subsection{Results}
\begin{figure*}[t]
\centering
\begin{minipage}{0.244\hsize}
   \includegraphics[width=\linewidth]{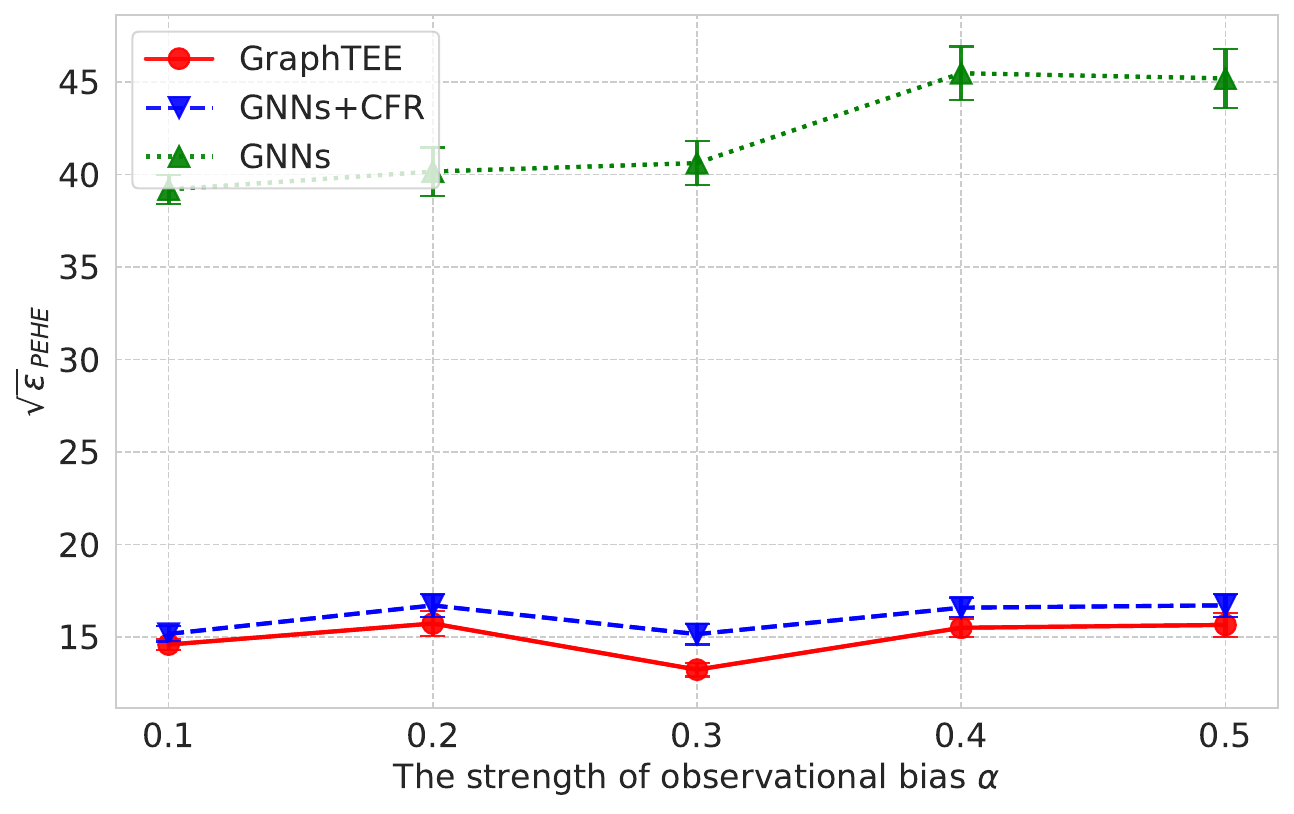}
      \centering
   (a): $\sqrt{\epsilon_{\text{PEHE}}}$~(Synthetic)
 \end{minipage}
 \begin{minipage}{0.244\hsize}
   \includegraphics[width=\linewidth]{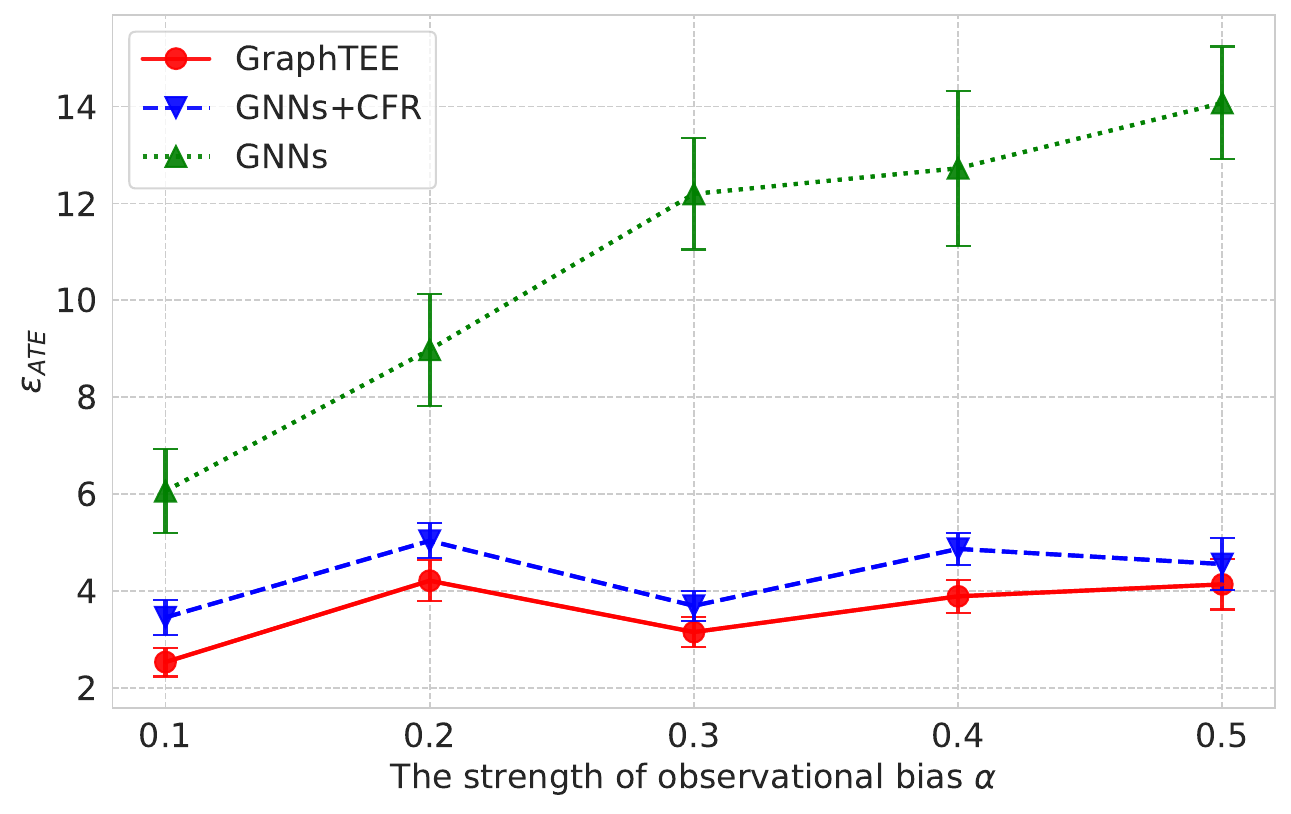}
      \centering
   (b): ${\epsilon_{\text{ATE}}}$~(Synthetic)
 \end{minipage}
 \begin{minipage}{0.244\hsize}
  \includegraphics[width=\linewidth]{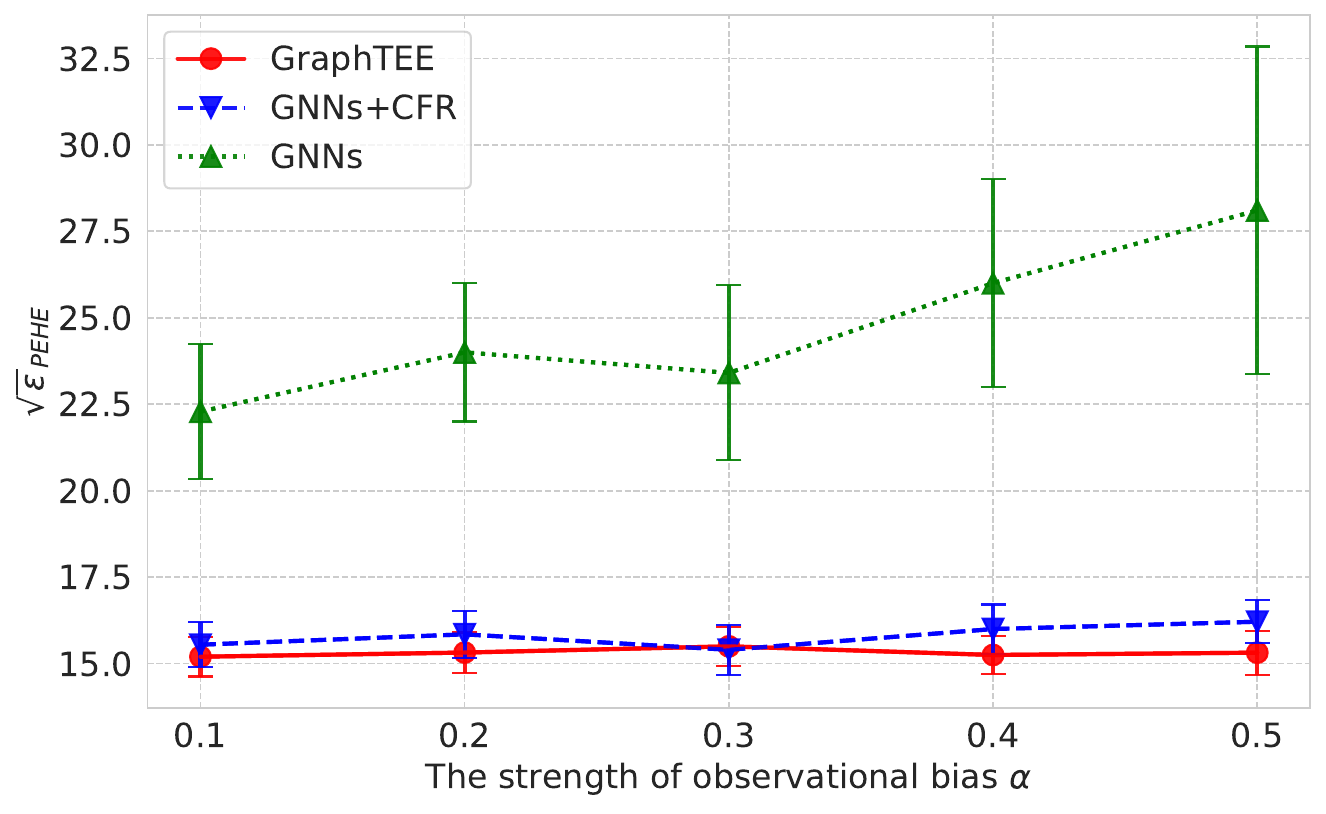}
      \centering
  (c): $\sqrt{\epsilon_{\text{PEHE}}}$~(Reddit)
 \end{minipage}
 \centering
  \begin{minipage}{0.244\hsize}
  \includegraphics[width=\linewidth]{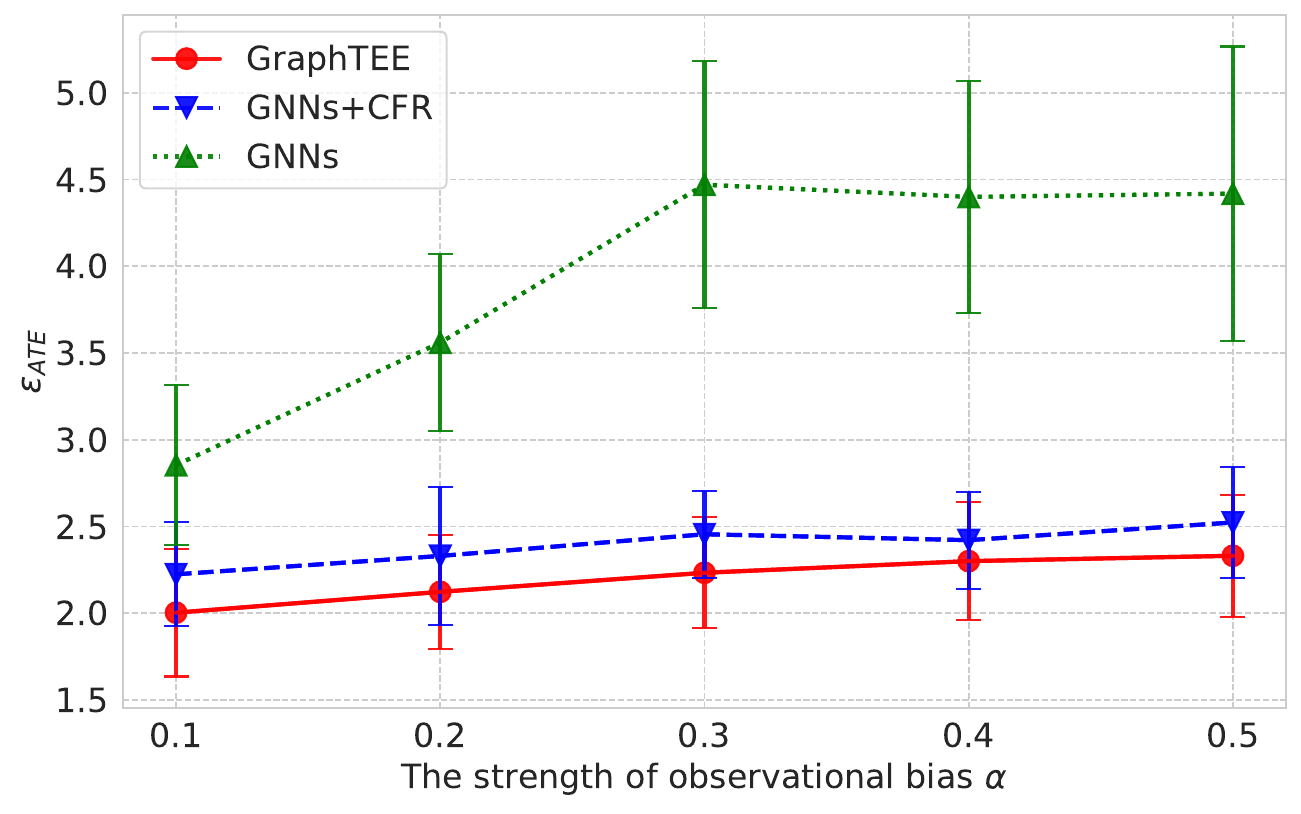}
      \centering
  (d): ${\epsilon_{\text{ATE}}}$~(Reddit)
 \end{minipage}\\
 \caption{{Sensitivity of the results to the strength of observational bias $\alpha$. The proposed method shows the robustness against the observational bias and consistently perform better than the baseline methods. Lower values are better. 
 }}\label{fig:bias}
\end{figure*}

   Table~\ref{table:results} summarizes the entire results in comparison with several baseline methods. 
   First, whereas mean or the DeepSets-based methods do not work well for the both datasets, methods making use of GNNs outperform these baseline methods significantly.
   Second, the bias mitigation techniques remarkably improve the predictive performances for the both datasets compared to GNNs without any such technique.
   The naive methods without any bias mitigation technique suffer severe observational bias and fail to predict treatment effect.
Conversely, GraphTEE significantly outperforms all the baseline methods in both the datasets. This result shows the effectiveness of efficient bias regularization and GraphTEE could mitigate bias.
Next, we investigate how the models perform against the strength of observational bias.
Figure~\ref{fig:bias} shows the performance comparisons when we change the strength of observational bias. GraphTEE consistently outperforms the baseline methods even when strong bias is given. Results of other ablation studies such as sensitivity to the strength of hyper-parameter are included in Appendix.
Figure~\ref{fig:reg} demonstrates the sensitivity to the strength of regularization. Although we need to somewhat exercise discretion in selecting to give the best performance, GraphTEE still performs better than the baseline methods.
To investigate the effect of the ratio of confounding nodes on the performance, we keep confounding nodes as nodes which have the highest degree in graphs and change the sizes of graphs, i.e., the number of nodes in a graph, for the synthetic dataset.

\begin{figure*}[th]
\centering
\begin{minipage}{0.245\hsize}
   \includegraphics[width=\linewidth]{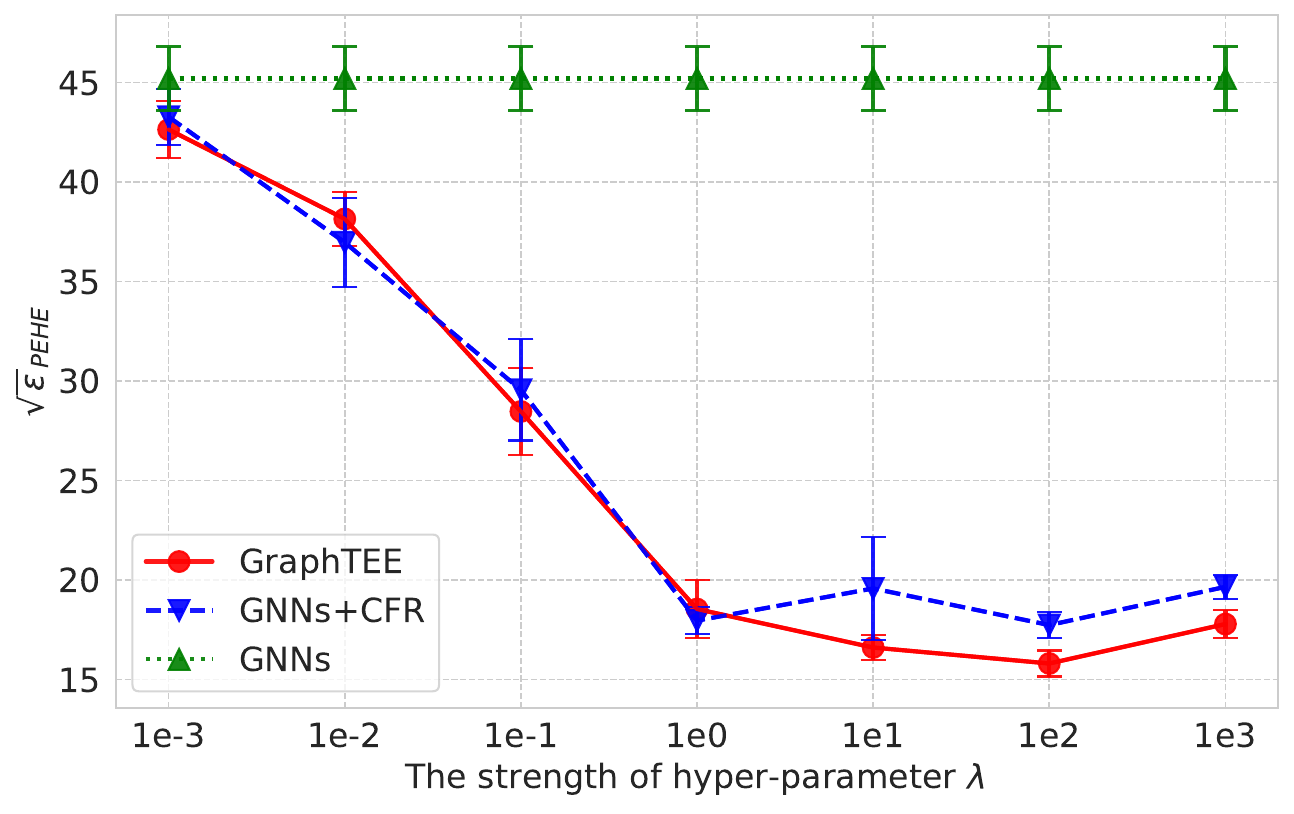}
      \centering
   (a): $\sqrt{\epsilon_{\text{PEHE}}}$~(Synthetic)
 \end{minipage}
 \begin{minipage}{0.245\hsize}
   \includegraphics[width=\linewidth]{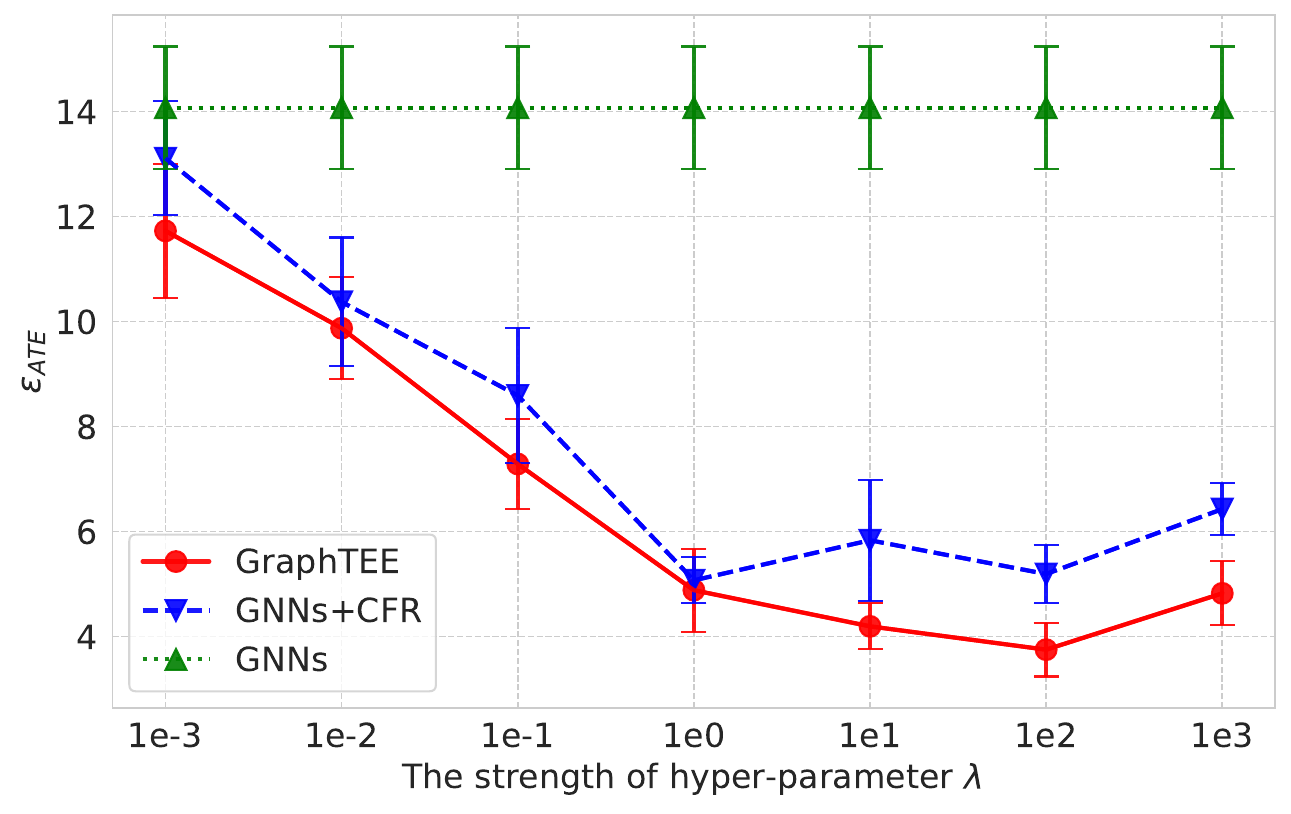}
      \centering
   (b): ${\epsilon_{\text{ATE}}}$~(Synthetic)
 \end{minipage}
 \begin{minipage}{0.245\hsize}
  \includegraphics[width=\linewidth]{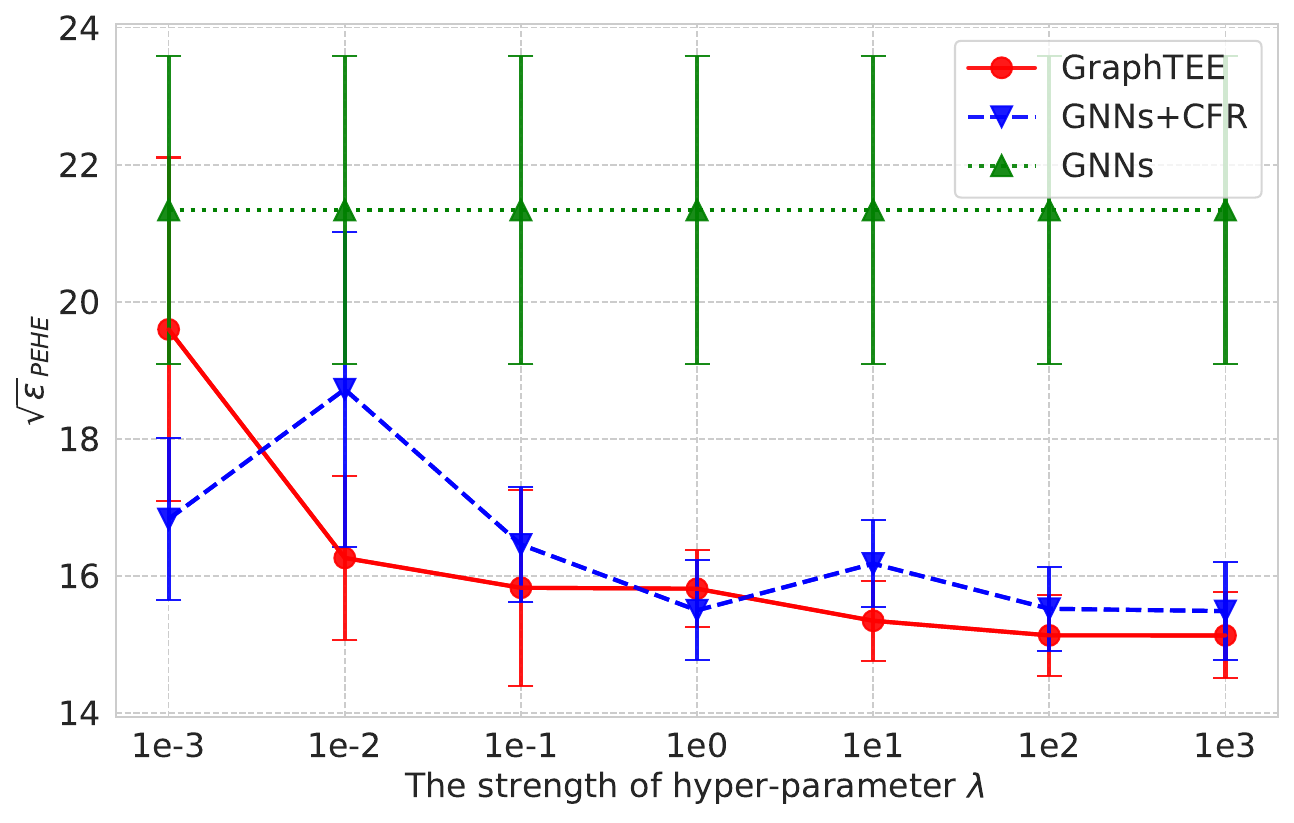}
      \centering
  (c): $\sqrt{\epsilon_{\text{PEHE}}}$~(Reddit)
 \end{minipage}
 \centering
  \begin{minipage}{0.245\hsize}
  \includegraphics[width=\linewidth]{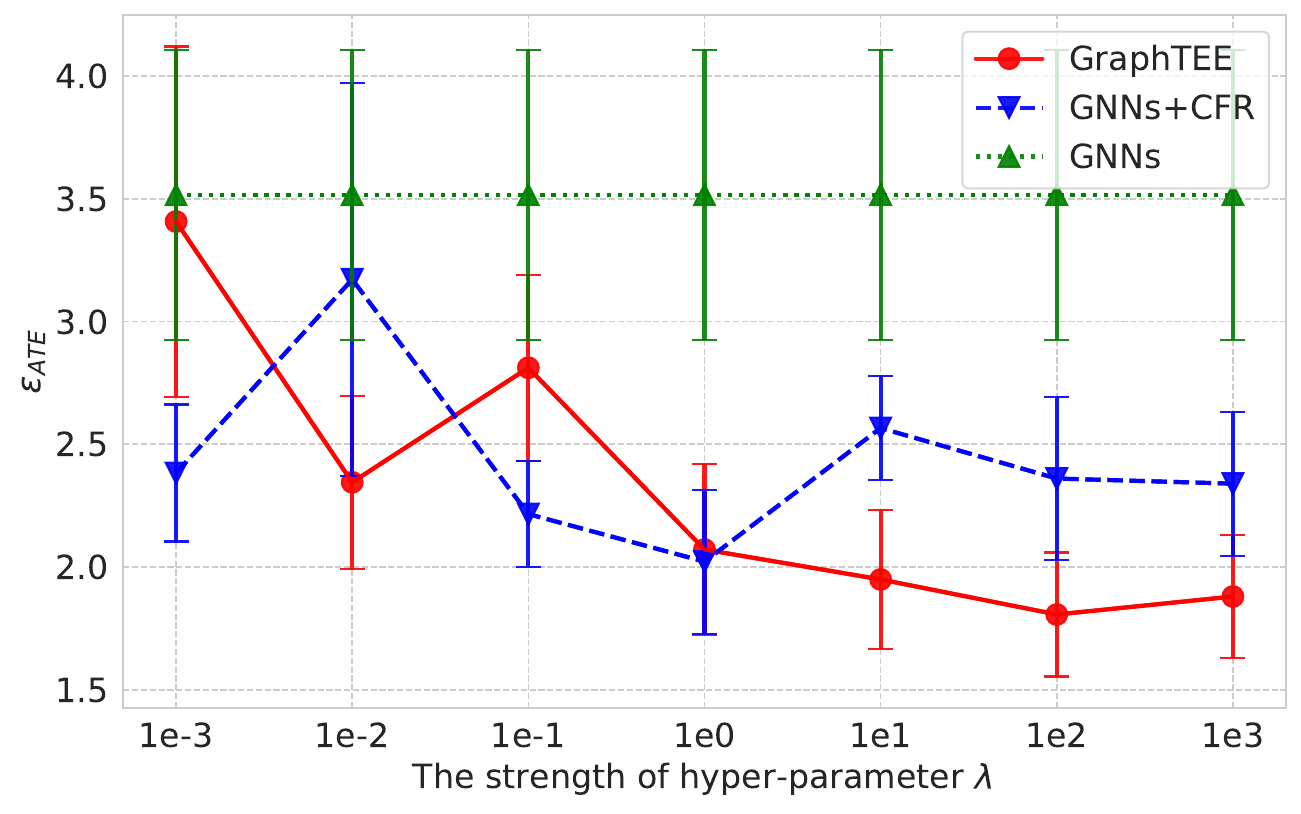}
      \centering
  (d): ${\epsilon_{\text{ATE}}}$~(Reddit)
 \end{minipage}\\
 \caption{{Sensitivity of the results to the strength of hyper-parameter $\lambda$. Although we need to exercise discretion in selecting the hyper-parameter, the proposed method consistently demonstrates the better results than the baseline methods. Lower values are better.
 }}\label{fig:reg}
\end{figure*}

\section{Conclusion}
In this study, we considered treatment effect estimation on graph-structured targets, which have a wide range of real world applications in different fields. We proposed GraphTEE, a two-step novel approach that aims to mitigate bias more efficiently. In the first step, we identify confounding nodes which have effects on treatment assignment. In the second step, we employ GNNs to predict outcomes while we perform bias mitigation by using confounding nodes. We also provide theoretical analyses showing how the proposed method can reduce the effect of observational bias more efficiently than existing approaches. In experiments using synthetic and semi-synthetic datasets, the proposed method outperformed baseline methods, especially with large-sized input graphs.
In this study, we mainly considered $1$-Lipschitz function space. 
We can not claim similar results as Theorem~\ref{theorem:main} currently for other function spaces and this might limit the range of application to some extent. Future work includes theoretical analyses towards that direction.

\bibliographystyle{tmlr}
\bibliography{main}

\clearpage
\appendix
\section{Proof}
\begin{lemma}\label{lemma:cfr}
Let $\Phi:\mc{G}\rightarrow\mc{Z}$ be a one-to-one mapping function.
Let $h\colon\mc{Z}\times\{0,1\}\rightarrow\mc{Y}$ be a hypothesis.
Let $\mc{F}$ be a family of functions and $f\in\mc{F}\colon\mc{Z}\rightarrow\mc{Y}$.
Assume there exists a constant ${B_\Phi}$ such that $\frac{1}{B_\Phi}\int_{\mc{Y}} L(Y^{t}, h(\Phi(G), t))p(Y^{t}\mid G)p(G, t)d{Y^{t}}\in\mc{F}$. Then, we have
\begin{align}
\epsilon_{\mathrm{test}}(h,\Phi) \leq & \epsilon_{\mathrm{train}}(h, \Phi)
+ B_{\Phi}\cdot \text{IPM}_{\mc{F}}(p({z_{(t=0)}}), p((z_{(t=1)}))),
\end{align}
where $z_{(t=0)}$ and $z_{(t=1)}$ stand for representations of control and treatment group, respectively.
\end{lemma}
\proof{See the original papers~\cite{shalit2017estimating,johansson2018learning}. We also used the following inequality.
\begin{align}
&\text{IPM}_{\mc{F}}(p(x, t), p(x)p(t))\nonumber\\&=\underset{f\in\mc{F}}{\sup}\Bigl| \int_{\mc{X}} f(x, t=0) p(t=0) (p(x\mid t=0)-p(x))dx\nonumber \\& + \int_{\mc{X}} f(x, t=1) p(t=1) (p(x\mid t=1)-p(x))dx  \Bigr|\\
&\leq \underset{f\in\mc{F}}{\sup} \Bigl| \int_{\mc{X}} f(x, t=0) p(t=0)p(t=1) (p(x\mid t=0) - p(x\mid t=1))dx \Bigr| \nonumber\\
&+\underset{f\in\mc{F}}{\sup} \Bigl| \int_{\mc{X}} f(x, t=1) p(t=0)p(t=1) (p(x\mid t=1)
- p(x\mid t=0))dx \Bigr|\\
&= p(t=0)p(t=1) \text{IPM}_{\mc{F}} (p(x\mid t=0), p(x\mid t=1))\nonumber\\&+p(t=0)p(t=1) \text{IPM}_{\mc{F}} (p(x\mid t=0), p(x\mid t=1))\\
&\leq \frac{1}{2}\text{IPM}_{\mc{F}}( p(x \mid t=0), p(x \mid t=1)) \quad(\because p(t=0)p(t=1)\leq\frac{1}{4})\\
&\leq \text{IPM}_{\mc{F}}( p(x \mid t=0), p(x \mid t=1)   ).
\end{align}}

\baselineproposition*
\proof{Combining Proposition~\ref{proposition:ipm_error} and Lemma~\ref{lemma:cfr}, we have the result.}
\begin{lemma}\label{lemma:hojo}
    Suppose we have the following error:
\begin{align}
    \text{IPM}_{\mc{F}_{ty}}(p(t)p(z_c, z_y), p(t)p(z_c)p(z_y))
    =\text{IPM}_{\mc{F}_y}(p(z_c, z_y), p(z_c)p(z_y))
    =\Delta,\end{align} then the following inequality holds.
    \begin{align}
    \text{IPM}_{\mc{F}_{ty}}(p(t\mid z_c)p(z_c, z_y), p(t)p(z_c)p(z_y))\leq \text{IPM}_{\mc{F}_{ty}}(p(t\mid z_c)p(z_c)p(z_y), p(t)p(z_c)p(z_y)) + 2\Delta.
    \end{align}
\proof{
We can rewrite $p(z_c, z_y)$ as $p(z_c, z_y)=p(z_c)p(z_y)+\Delta_{z_c, z_y}$ using a difference $\Delta_{z_c, z_y}$ between $p(z_c, z_y)$ and $p(z_c)p(z_y)$ at $z_c$ and $z_y$.
Then, we can derive the following equation.
\begin{align}
&\text{IPM}_{\mc{F}_{y}}(p(z_c, z_y), p(z_c)p(z_y))\nonumber\\&=
\underset{f\in\mc{F}_{y}}{\sup}\Bigl|(\int_{\mc{Z}_c\times{\mc{Z}_y}}f(z_c, z_y)\Bigl( p(z_c, z_y) - p(z_c)p(z_y)\Bigr)dz_{c}dz_{y}\Bigr|\\
&=\underset{f\in\mc{F}_{y}}{\sup}\Bigl| (\int_{\mc{Z}_c\times{\mc{Z}_y}}f(z_c, z_y)\Bigl( (p(z_c)p(z_y)+\Delta_{z_c, z_y} - p({z_c})p({z_y})\Bigr)dz_{c}dz_{y}\Bigr|\\
&=\underset{f\in\mc{F}_{y}}{\sup}\Bigl|(\int_{\mc{Z}_c\times{\mc{Z}_y}}f(z_c, z_y)\Delta_{{z_c, z_y}dz_{c}dz_y}\Bigr|\\
&=\Delta.\label{eq:delta}
\end{align}
Then, the claim is proved as follows. 
\begin{align}
&\text{IPM}_{\mc{F}_{ty}}(p(t\mid z_c)p({z_c, z_y}), p(t)p(z_c)p(z_y)))\\
&=\underset{f\in\mc{F}_{ty}}{\sup}\Bigl|(\int_{\mc{T}\times\mc{Z}_c\times{\mc{Z}_y}}f(t, z_c, z_y)\Bigl( p(t\mid z_c)p(z_c, z_y) -p(t)p(z_c)p(z_y)\Bigr)d{tdz_cdz_y}\Bigr|\\
&=\underset{f\in\mc{F}_{ty}}{\sup}\Bigl|(\int_{ \mc{T}\times{\mc{Z}_c}\times{{\mc{Z}_y}}}f(t, {z_c, z_y})\Bigl( p(t\mid {z_c})(p({z_c})p({z_y}) + \Delta_{z_c, z_y})- p(t)p({z_c})p({z_y})\Bigr)dtd{z_c}d{z_y}\Bigr|\\
&\leq\text{IPM}_{\mc{F}_{ty}}(p(t\mid {z_c})p({z_c})p({z_y}), p(t)p({z_c})p({z_y})) + \underset{f\in\mc{F}_{ty}}{\sup}\Bigl|\int_{{\mc{T}}\times{\mc{Z}_c}\times{{\mc{Z}_y}}}f(t, {z_c, z_y}) p(t\mid{z_c})\Delta_{{z_c, z_y}}d{td{z_c}d{z_y}}\Bigr|
\ \  \\&\quad(\because\text{triangle inequality})\nonumber\\
&\leq\text{IPM}_{\mc{F}_{ty}}(p(t\mid {z_c})p({z_c})p({z_y}), p(t)p({z_c})p({z_y}))+ 2\Delta. \\ &\ \ (\because \text{Eq.}~(\ref{eq:delta}),\ 0<p(t)<1,\ 0<p(t\mid z_c)<1 ).\nonumber
\end{align}
}
\end{lemma}

\maintheorem*
\proof{
Using Lemma~\ref{lemma:hojo}, we have
\begin{align}
    \text{IPM}_{\mc{F}_{ty}}(p(t\mid z_c)p(z_c, z_y), p(t)p(z_c)p(z_y))\leq \text{IPM}_{\mc{F}_{ty}}(p(t\mid z_c)p(z_c)p(z_y), p(t)p(z_c)p(z_y)) + 2\Delta.
\end{align}
Therefore,
\begin{align}
&\text{IPM}_{\mc{F}_{ty}}(p(t\mid {z_c})p({z_c, z_y}), p(t)p({z_c, z_y}) )\nonumber\\
&\leq \text{IPM}_{\mc{F}_{ty}}(p(t\mid {z_c})p({z_c, z_y}), p(t)p({z_c})p({z_y})) ) + \text{IPM}_{\mc{F}_{ty}}(p(t)p({z_c, z_y}), p(t)p({z_c})p({z_y}))\\
&\leq \text{IPM}_{\mc{F}_{ty}}(p(t\mid {z_c})p({z_c})p({z_y}), p(t)p({z_c})p({z_y})) + 2\Delta\nonumber
\\&+\text{IPM}_{\mc{F}_{ty}}(p(t)p({z_c, z_y}), p(t)p({z_c})p({z_y}))\\
&=\text{IPM}_{\mc{F}_c}(p(t\mid {z_c})p({z_c}), p(t)p({z_c}) ) + 2\Delta+{\underbrace{\text{IPM}_{\mc{F}_y}(p({z_c, z_y}), p({z_c})p({z_y}))}_{=\Delta}}.
\end{align}
Using the theoretical analysis in ~\cite{sriperumbudur2009integral},
at least probability 1-$\delta$,
\begin{align}
&\text{IPM}_{\mc{F}}(p(t\mid {z_c})p({z_c, z_y}), p(t)p({z_c, z_y}) )-\text{IPM}_{\mc{F}}(\hat{p}(t\mid {z_c})\hat{p}({z_c, z_y}), \hat{p}(t)\hat{p}({z_c, z_y}) )\nonumber\\
&\leq 2R_{m}(\mc{F})+2R_{n}(\mc{F})+\sqrt{{18\nu^{2}}\log \frac{4}{\delta}}({\frac{1}{\sqrt{m}}}+{\frac{1}{\sqrt{n}}})
\end{align}
Here, we claim there exists a constant $C_{c}~(1\geq C_{c}>0)$ such that
\begin{align}
C_{c}\sup f({z_c, z_y,},t)=\sup g({z_c},t).
\end{align}
This is because
\begin{align}
 \forall f\in\mc{F}, \sup g({z_c}, t)&=\sup \mathbb{E}_{{z_y}} [f({z_c, z_y,},t)]\\
 &\leq \mathbb E_{{z_y}}[{\sup f({z_c, z_y}, t)}]\\
 &\leq {\sup f({z_c, z_y}, t)}.
\end{align}
Therefore, we focus on an empirical error with regard to the following IPM.
\begin{align}
&\text{IPM}_{\mc{F}}(p({z_{c(t=1)}}), p({z_{c(t=0)}})  ) - \text{IPM}_{\mc{F}}(
 \hat{p}({z_{c(t=1)}}), \hat{p}({z_{c(t=0)}}  )).
\end{align}
By performing similar operation, we have with probability at least $1- \frac{\delta}{8}$,
\begin{align}
\underset{f\in\mc{F}}{\sup}\Bigl| \mathbb{P}_mf-Pf \Bigr|&\leq \mathbb{E} \underset{f\in\mc{F}}{\sup}\Bigl| \mathbb{P}_mf-Pf \Bigr|+\sqrt{\frac{2C_{c}^{2}\nu^{2}} {m}    \log \frac{8}{\delta}} \\
&\leq 2\mathbb{E}\underset{f\in\mc{F}}{\sup}\Bigl|\frac{1}{m}\sum^{m}_{i=1}\sigma_if({z_c}, t) \Bigr| + \sqrt{\frac{2C_{c}^{2}\nu^{2}} {m}    \log \frac{8}{\delta}}.
\end{align}
we have with probability at least  $1-\frac{\delta}{8}$,
\begin{align}
\mathbb{E}\underset{f\in\mc{F}}{\sup}\Bigl|\frac{1}{m}\sum^{m}_{i=1}\sigma_if({z_c}, t) \Bigr|\leq \mathbb{E}_{\sigma} \underset{f\in\mc{F}}{\sup}\Bigl|\frac{1}{m}\sum^{m}_{i=1}\sigma_if({z_c}, t) \Bigr| +\sqrt{\frac{2C_{c}^{2}\nu^{2}} {m}    \log \frac{8}{\delta}}.
\end{align}
we have that with probability at least $1-\frac{\delta}{4}$
\begin{align}
\underset{f\in\mc{F}}{\sup}\Bigl| \mathbb{P}_mf-Pf \Bigr|
&\leq 2R_{m}(\mc{F}) + \sqrt{\frac{18C_{c}^{2}\nu^{2}} {m}    \log \frac{8}{\delta}}.
\end{align}
Then we have that with probability at least $1-\frac{\delta}{2}$,
\begin{align}
&\Bigl|\text{IPM}_{\mc{F}_c}(p({z_{c(t=1)}}), p({z_{c\mid t=0}})  ) - \text{IPM}_{\mc{F}_c}(\hat{p}({z_{c(t=1)}}), \hat{p}({z_{c(t=0)}})  )\Bigr|\nonumber\\
&\leq  2R_{m}(\mc{F}_c) +2R_{n}(\mc{F}_c) + \sqrt{\frac{18C_{c}^{2}\nu^{2}} {m}    \log \frac{8}{\delta}} + \sqrt{\frac{18C_{c}^{2}\nu^{2}} {n}    \log \frac{8}{\delta}}.
\end{align}
Next we perform similar analysis on $\text{IPM}_{\mc{F}}(p(t)p({z_c, z_y}), p(t)p({z_c})p({z_y}))$.
Then we have that with probability at least $1-\frac{\delta}{2}$,
\begin{align}
&\Bigl|\text{IPM}_{\mc{F}_y}(p({z_c, z_y}), p({z_c})p({z_y})) - \text{IPM}_{\mc{F}_y}(\hat{p}({z_c, z_y}), \hat{p}({z_c})p({z_y}))\Bigr|\nonumber\\
&\leq  2R_{m+n}(\mc{F}_y) +2R_{m+n}(\mc{F}_y) + \sqrt{\frac{18\nu^{2}} {m+n}    \log \frac{8}{\delta}} + \sqrt{\frac{18\nu^{2}} {m+n}    \log \frac{8}{\delta}}.
\end{align}
Finally, by combining these results, we have with probability at least $1-{\delta}$,
\begin{align}
&\Bigl|\text{IPM}_{\mc{F}_c}(p({z_{c(t=1)}}), p({z_{c(t=0)}}  )) - \text{IPM}_{\mc{F}_c}(\hat{p}({z_{c(t=1)}}), \hat{p}({z_{c(t=0)}})  )\Bigr|\nonumber\\+&\Bigl|\text{IPM}_{\mc{F}_y}(p({z_c, z_y}), p({z_c})p({z_y}))-\text{IPM}_{\mc{F}_y}(\hat{p}({z_c, z_y}), \hat{p}({z_c})\hat{p}({z_y}))\Bigr|\nonumber\\
\leq &2R_{m}(\mc{F}_c) +2R_{n}(\mc{F}_c) + \sqrt{{18C_{c}^{2}\nu^{2}}    \log \frac{8}{\delta}} (\frac{1}{\sqrt{m}}+\frac{1}{\sqrt{n}})\nonumber
\\+& 2R_{m+n}(\mc{F}_y) +2R_{m+n}(\mc{F}_y) + 2\sqrt{\frac{18\nu^{2}}{m+n}\log\frac{8}{\delta}}
\\\leq &2R_{m}(\mc{F}_c) +2R_{n}(\mc{F}_c) + \sqrt{{18C_{c}^{2}\nu^{2}}    \log \frac{8}{\delta}} (\frac{1}{\sqrt{m}}+\frac{1}{\sqrt{n}})\nonumber\\
+&4R_{m+n}(\mc{F}_y) +
2\sqrt{\frac{18\nu^{2}}{m+n}\log\frac{8}{\delta}}.
\end{align}

\end{document}